\documentclass[10pt,twocolumn,letterpaper]{article}

\usepackage[pagenumbers]{cvpr} 
\usepackage[dvipsnames]{xcolor}
\usepackage{multirow}
\usepackage{times}
\usepackage{latexsym}

\usepackage[T1]{fontenc}

\usepackage[utf8]{inputenc}

\usepackage{microtype}

\usepackage{inconsolata}

\usepackage{comment}
\usepackage{ifthen} 
\usepackage{overpic}
\usepackage{booktabs}
\usepackage{multirow}
\usepackage{rotating}
\usepackage{wrapfig}
\usepackage{amsmath}
\usepackage{amssymb}
\usepackage{bbm}
\usepackage{dsfont}
\usepackage{wrapfig}
\usepackage{verbatim} 
\usepackage{nicefrac} 
\usepackage{caption}
\usepackage{pifont}
\usepackage{colortbl} 
\usepackage{color}
\usepackage{xcolor}
\usepackage{balance}
\usepackage{marvosym}

\usepackage{float}

\usepackage{bbding}

\definecolor{cvprblue}{rgb}{0.21,0.49,0.74}
\usepackage[pagebackref,breaklinks,colorlinks,citecolor=cvprblue]{hyperref}

\def\paperID{*****} 
\def\confName{CVPR}
\def\confYear{2024}

\title{Lenna: Language Enhanced Reasoning Detection Assistant}

\author{Fei Wei$^{1}$, Xinyu Zhang$^{1}$, Ailing Zhang$^{1,2}$, Bo Zhang$^{1}$, Xiangxiang Chu$^{1}$\footnotemark[1]
\\
{\small $^{1}$ Meituan Inc.}
{\small $^{2}$ Beihang University}
\\
{\tt\small\{weifei06, zhangxinyu35\}@meituan.com}
{\tt\small\{annyzhang\}@buaa.edu.cn}
{\tt\small\{zhangbo97, chuxiangxiang\}@meituan.com}
}

\begin{document}
\maketitle
\footnotetext[1]{Corresponding author. }
\footnotetext[2]{This work is done when Ailing Zhang is an intern at Meituan. This project is still under development and the reported results are subject to frequent modifications.}
\begin{abstract}
With the fast-paced development of multimodal large language models (MLLMs), we can now converse with AI systems in natural languages to understand images. However, the reasoning power and world knowledge embedded in the large language models have been much less investigated and exploited for image perception tasks. 
In this paper, we propose \textbf{Lenna}, a \textbf{L}anguage \textbf{e}nhanced reaso\textbf{n}ing detectio\textbf{n} \textbf{a}ssistant, which utilizes the robust multimodal feature representation of MLLMs, while preserving location information for detection. This is achieved by incorporating an additional \texttt{<DET>} token in the MLLM vocabulary that is free of explicit semantic context but serves as a prompt for the detector to identify the corresponding position. 
To evaluate the reasoning capability of Lenna, we construct a ReasonDet dataset to measure its performance on reasoning-based detection. Remarkably, Lenna demonstrates outstanding performance on ReasonDet and comes with significantly low training costs. It also incurs minimal transferring overhead when extended to other tasks. Our code and model will be available at \url{https://github.com/Meituan-AutoML/Lenna}.
\end{abstract}  
\section{Introduction}
\begin{figure}[t]
  \centering
   \includegraphics[width=0.97\linewidth]{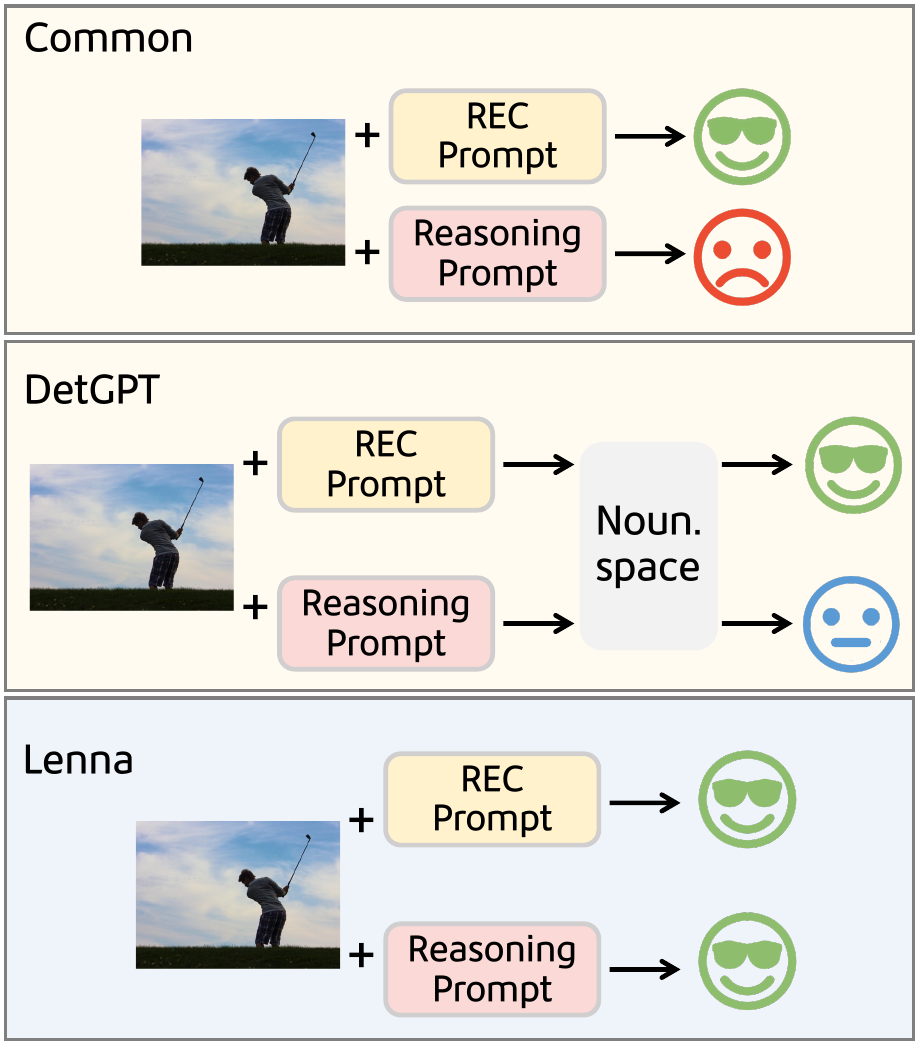}
   \caption{Illustration of common MLLM-based REC  methods, DetGPT, and Lenna. A REC prompt comprises explicit instructions of the target, typically describing its position, color, and shape (\eg. “\emph{A long, slender golf club}”). A reasoning prompt contains implicit intentions (\eg. “\emph{something showing that the man is playing sports}”). Common methods fail to handle the reasoning prompt effectively. DetGPT struggles to translate the input prompts to the target names in the noun space. Lenna excels at handling both.}
   \label{fig:fig1}
\end{figure}

\label{sec:intro}
Recent accelerating advancement of large language models (LLM)~\citep{du2022glm, touvron2023llama2, scao2022bloom} has amplified the model's capacity for natural language comprehension and generation. Bolstered by these large language models, multimodal large language models (MLLM) have achieved significant performance leaps in perception tasks (\eg, detection~\citep{wang2023visionllm, pi2023detgpt}, segmentation~\citep{lai2023lisa}), generation tasks (\eg, captioning~\citep{li2023blip2}, VQA~\citep{zhu2023minigpt, liu2023llava}).

Referring Expression Comprehension (REC), serving as a crucial task for assessing the natural language understanding and positioning capability of multimodal large models, has been the focal point of numerous studies, \eg, InstructDet~\citep{dang2023instructdet}, Shikra~\citep{chen2023shikra}, miniGPT-v2~\citep{chen2023minigpt}. The objective of the REC task is to obtain the position of an object given explicit instructions, such as positional words, colors, shapes, \etc. Nonetheless, rare attention has been paid to the model's reasoning ability which demands an understanding of implicit intentions. DetGPT~\citep{pi2023detgpt} represents the inaugural research to propose the reasoning-based object detection task. Yet its weak connection of the reasoning and the localization process, via an intermediate bridge in the form of several nouns, results in substantial information loss in the multimodal feature space. Figure~\ref{fig:fig1} illustrates the differences among the common MLLM-based REC methods, DetGPT, and our approach.

To facilitate the reseasoning capacity and world knowledge of LLMs in the perception task, we introduce Lenna, a \textbf{L}anguage \textbf{E}nhanced reaso\textbf{N}ing detectio\textbf{N} \textbf{A}ssistant. Drawing inspiration from LISA~\cite{lai2023lisa}, we discover that the token embedding of LLMs possesses a robust multimodal feature representation capability, which can transmit localization information to the detector without forfeiting reasoning ability. To achieve this, we incorporate an additional \texttt{<DET>} token as a signal to convey object detection information. Unlike other words in the existing vocabulary, \texttt{<DET>} token is free of explicit semantic context. The hidden embedding of the \texttt{<DET>} token functions as a prompt for the detector, assisting it in pinpointing the relevant target position.

Owing to the simplistic design of the \texttt{<DET>} token, we broaden the REC-based object detection to reasoning-based object detection without altering their original design. We curate a ReasonDet dataset processed from ReasonSeg~\citep{lai2023lisa} to evaluate reasoning detection capacity. Leveraging LLM's world knowledge and intelligence, Lenna also exhibits commendable performance on ReasonDet.

It is noteworthy that in contrast to other existing works, Lenna incurs a significantly lower training cost. We expended merely 20 hours in training on 8 A100 GPUs. Due to the streamlined design, Lenna also holds the potential to be extended to other tasks (\eg, instance segmentation, grounding) with minimal cost implications.

Our main contributions can be summarized as follows.

\begin{itemize}
\item
We propose Lenna, our language-enhanced reasoning detection assistant, that incorporates REC-based and reasoning-based detection in the same simplistic and extensible framework. 
\item 
We curate a benchmark dataset called ReasonDet to quantitatively measure the reasoning detection performance of MLLMs.
\item Lenna comes with inexpensive training cost and outperforms previous MLLMs on REC and ReasonDet. Concurrently, the visualization results from ReasonDet affirm Lenna's consistent capability in reasoning object detection.
\end{itemize}

\begin{figure*}[t]
\begin{center}
\includegraphics[width=0.97\textwidth]{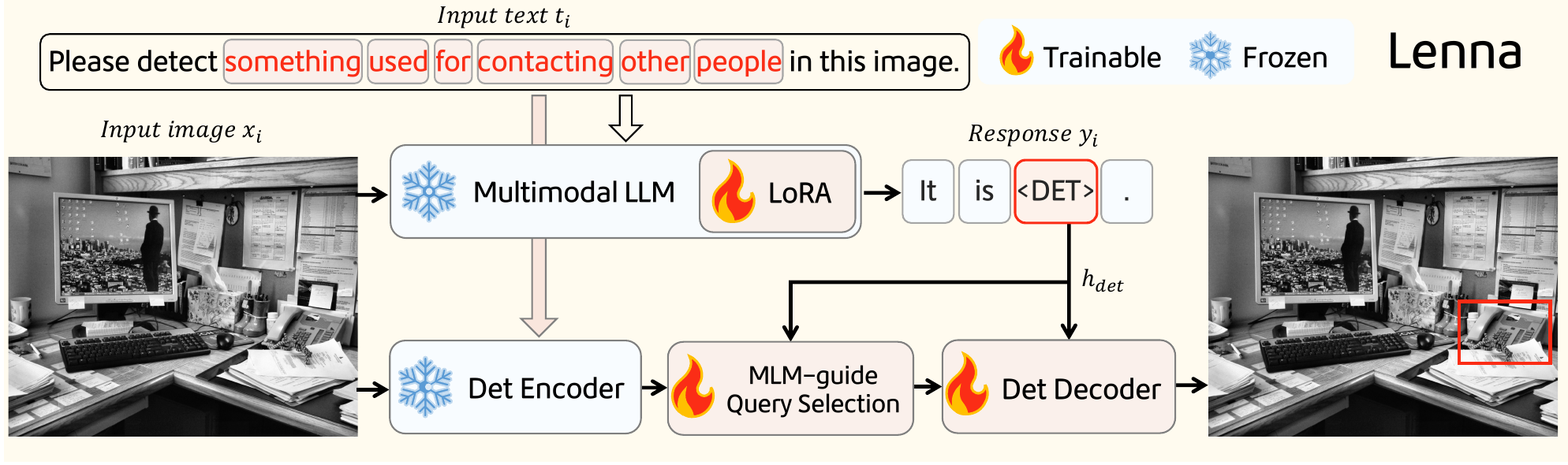}
\end{center}
   \caption{The pipeline of Lenna. Given an input image $x_i$ and text $t_i$, the multimodal LLM generates a text output. The last-layer embedding for the \texttt{<DET>} token is subsequently incorporated into $\mathsf{MLM\text{-}guide Query Selection}$ and $\mathsf{Det Decoder}$ to ascertain the object's location. The $\mathsf{Det}$ model is architecturally based on Grounding-DINO, which takes text caption $t_i'$ (marked in red) out of $t_i$ and the image $x_i$ as input.}
\label{fig:lenna}
\end{figure*}

\section{Related Work}
\label{sec:related}

\subsection{Referring Expression Comprehension}
The task of Referring Expression Comprehension (REC) is aimed at locating the objects that are explicitly referred to by free-form guided language expressions. This task has been widely adopted by a plethora of works~\citep{dang2023instructdet, chen2023shikra, wang2023cogvlm, yang2022unitab} as a standard for monitoring the performance of their models. The evaluation covers language understanding and object localization, offering a holistic measurement of the model's competencies.
GLIP~\citep{li2021grounded} unifies object detection and grounding by reformulating object detection as a phrase grounding task. 
Grounding-DINO~\citep{liu2023gdino} extends an original detector DINO~\citep{zhang2022dino} to an open-set detector by performing vision-language modality fusion with BERT~\citep{devlin2018bert} in multiple phases. 
DQ-DETR~\citep{liu2023dq} utilizes dual decoupled queries to alleviate the difficulty of modality alignment between image and text in the DETR~\citep{carion2020detr} framework.
Moreover, benefiting from the development of language models, a lot of works~\citep{chen2021pix2seq, wang2022ofa, zou2022xdecoder, cheng2021mask2former, lu2022unified, wang2022git} cast object detection as a language modeling task and achieve better performance on REC.
PEVL~\citep{yao2022pevl} reformulates discretized object positions and language in a unified language modeling framework.
UniTab~\citep{yang2022unitab} considers texts and box predictions as an auto-regressive token generation task and presents a unified encoder-decoder model fully shared for texts, boxes, and alignment predictions.
However, in scenarios where the text lacks an explicit referent and necessitates reasoning from the language model, current models are largely constrained by the comprehension capacity, thereby inadequately addressing such situations. Conversely, Lenna, schemed to leverage the benefits of LLM, exhibits proficiency in understanding complex semantics, reasoning in the context, and accurately pinpointing the target content.

\subsection{Multimodal Large Language Model}
Large Language Models (LLMs)~\citep{zeng2022glm, touvron2023llama2, scao2022bloom, driess2023palm, du2022glm} have exceptional performance in diverse natural language processing (NLP) tasks. Expanding large language models to a multimodal version~\citep{wang2021simvlm, li2023otter, kosmos, ye2023mplug} has naturally obtained increasing attention. The exemplary efficacy of incumbent Vision Transformers~\citep{liu2021swin, dosovitskiy2020image, radford2021learning, chu2021Twins, pmlr-v139-touvron21a, chu2023CPVT} lays a robust groundwork for the expansion.
Flamingo~\citep{alayrac2022flamingo} employs adaption layers to pretrained LLM layers to incorporate visual information for the next-token prediction task.
LLaVA~\citep{liu2023llava} makes the first attempt to utilize GPT4~\citep{openai2023gpt4} to generate visual instruction data and converts image feature into the word embedding space with a simple linear layer.
Shikra~\citep{chen2023shikra} introduces the task of Referential Dialogue and injects visual grounding capabilities into LLMs.
MiniGPT-v2~\citep{chen2023minigpt} proposes a unified interface model for handling various vision-language tasks by a task-oriented instruction training scheme.
BLIP-2~\citep{li2023blip2} designs a Q-Former module to feed the most useful visual feature to the LLM to have the desired text output.
VisionLLM~\citep{wang2023visionllm} aligns the definitions of vision-centric tasks with the methodologies of LLMs.
Closely related to us, DetGPT\citep{pi2023detgpt} represents an initial foray into reasoning object detection, establishing a linkage between MLLM and open-vocabulary detector through the medium of object names. Nonetheless, the reliance on object names substantially curtails the representational prowess of the feature space. In stark contrast, Lenna exploits the rich characterization inherent in MLLM embeddings, which encapsulate and transmit semantic and localization information to the detector.




\section{Method}
\label{sec:method}
\subsection{Architecture Design}
\begin{figure}[t]
\begin{center}
\includegraphics[width=0.45\textwidth]{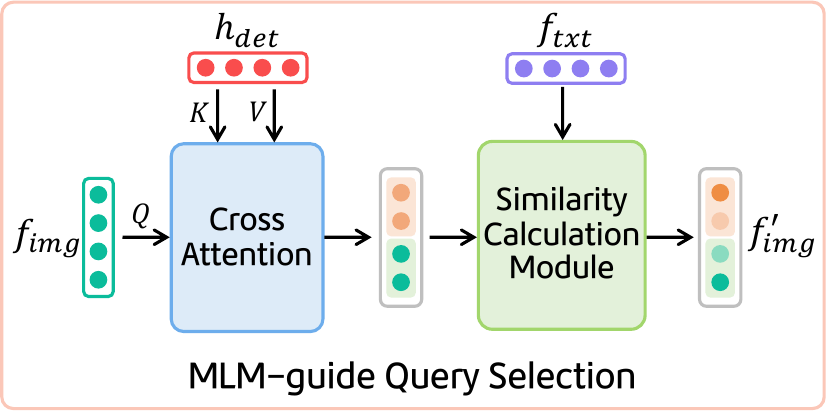}
\end{center}
   \caption{Architecture illustration of $\mathsf{MLM\text{-}guide Query Selection}$ module (MQS). The enhanced features $f_{img}$ and $f_{txt}$ are extracted from the encoder of the detector. $h_{det}$ represents the embedding corresponding to the \texttt{<DET>} token. The output of MQS is denoted as $f_{img}'$, which serves as the input feature of the decoder.}
\label{fig:MQS}
\end{figure}

We introduce Lenna, an end-to-end language-enhanced reasoning detection assistant. Figure~\ref{fig:lenna} delineates its framework. In essence, Lenna is an amalgamation of a multimodal large language model LLaVA~\citep{liu2023llava} and an open-set detector Grounding-DINO~\citep{liu2023gdino}. Analogous to LISA~\citep{lai2023lisa}, we initially extend the original LLM vocabulary with a special token \texttt{<DET>} to signify the demand for detection output. Upon receiving an image $x_{i}$ and a text instruction $t_{i}$, the multimodal large language model $\mathcal{M}$ engenders a text response $y_{i}$, which is formulated as
\begin{equation}
y_{i}=\mathcal{M}(x_{i},t_{i}).
\end{equation}
A \texttt{<DET>} token is necessitated to distinguish task types when MLLM is asked to undertake an object detection assignment. Consequently, we obtain an embedding $h_{det}$ that corresponds to the \texttt{<DET>} token, which is laden with both semantic and location information related to the target. Simultaneously, the pre-trained encoder of the detector, denoted as $\mathcal{D}_{enc}$, extracts enhanced image feature $f_{img}$ and text feature $f_{txt}$, which can be formulated as
\begin{equation}
f_{img}, f_{txt}=\mathcal{D}_{enc}(x_{i},t_{i}').
\end{equation}
where $t_{i}'$ is the object caption (marked in red in Fig.~\ref{fig:lenna}) in $t_{i}$. Subsequently, $h_{det}$, $f_{img}$ and $f_{txt}$ are fed into the MLM-guide query selection (MQS) module, which facilitates cross-space alignment between the feature spaces of BERT-based and LLM-based models. For better modality alignment in the implementation, MQS is designed to incorporate both a cross-attention module and a similarity calculation module, as shown in Figure~\ref{fig:MQS}. In the \textit{cross-attention module}, we employ $h_{det}$ as $K$, $V$ to activate the corresponding features in the enhanced image feature $f_{img}$. In the \textit{similarity calculation module}, akin to Grounding-DINO, we select features that exhibit greater relevance to the input text feature $f_{txt}$. The operation of MQS can be formulated as
\begin{equation}
f_{img}'=MQS(h_{det}, f_{img}, f_{txt})
\end{equation}

Ultimately, $h_{det}$ is incorporated into each text cross-attention layer of decoder $\mathcal{D}_{dec}$ resulting in the final location $pred$, which can be represented as

\begin{equation}
pred=\mathcal{D}_{dec}(h_{det}, f_{img}').
\end{equation}

\subsection{Optimization Objectives}
We construct an end-to-end training process where Lenna employs a loss function that combines the auto-regressive language modeling loss $\mathcal{L}_{tok}$ and the detection loss $\mathcal{L}_{det}$, amalgamated according to a specific weight ratio $\lambda_{tok}$ and $\lambda_{det}$, which can be mathematically expressed as 
\begin{equation}
\mathcal{L}=\lambda_{tok}\mathcal{L}_{tok}+\lambda_{det}\mathcal{L}_{det}
\end{equation}
In auto-regressive language modeling loss $\mathcal{L}_{tok}$, we maximize the likelihood of target token $y_t$ conditioned on input image $x_i$, input text $t_i$, and previous target tokens $y_{j'}$, which can be formulated as 
\begin{equation}
\mathcal{L}_{tok}= - \sum_{j=1}^{L}\log{p} \left [y_j|y_{j'}, x_i, t_i  \right ] \quad \text{where} \quad j' < j
\end{equation}
We follow Grounding-DINO in $\mathcal{L}_{det}$ to use the L1 Loss and the GIOU loss~\citep{rezatofighi2019generalized} as bounding box regression, and use contrastive loss~\citep{radford2021learning} for classification, which is formulated as 
\begin{equation}
\begin{split}
\mathcal{L}_{det}&=\lambda_{L1}L1(pred, gt)\\
&+\lambda_{GIOU}GIOU(pred, gt) \\
&+\lambda_{Contrast}Contrast(pred, gt)
\end{split}
\end{equation}

\begin{figure*}[t]
  \centering
   \includegraphics[width=0.8\linewidth]{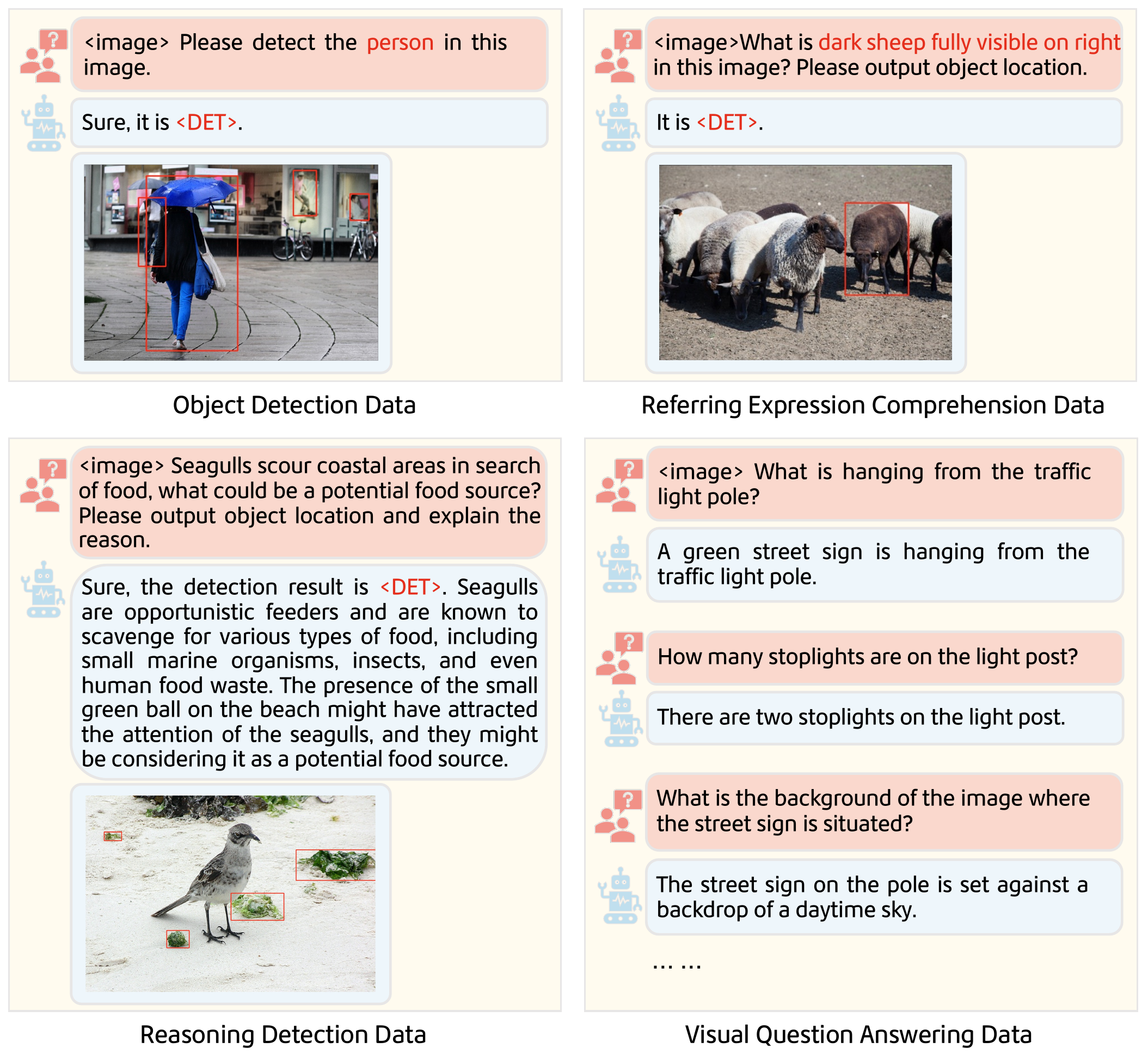}
   \caption{Examples of various types of training data for Lenna, including object detection data (OD), referring expression comprehension data (REC), reasoning detection data (RD), and visual question answering (VQA) data. In addition to the VQA data, the response of the remaining data incorporates the \texttt{<DET>} token, signifying the activation of the detection task. Correspondingly, the ground truth of its associated regression box is also present in red.}
   \label{fig:data}
\end{figure*}

\subsection{Training Data Formulation}
As shown in Figure~\ref{fig:data}, our model is trained on four distinct data types, which serve as a strategy to endow our model with image-text feature matching capabilities and proficiency by aligning two multimodal feature spaces. 
\paragraph{Object detection data.}
Typically, the ground truth data for object detection contains the positional and categorical information of all targets within a predefined list of categories. During the training phase, we formulate questions and answers adhering to a specific template:
\emph{“\textbf{User:} } \texttt{<image>} \emph{Please detect the }\{category\} \emph{in this image. \textbf{Assistant:} Sure, }\texttt{<DET>}\emph{.”},
wherein the \{category\} is randomly chosen from the ground truth categories present in the image, \texttt{<image>} is the placeholder of image tokens. The indicator \texttt{<DET>} denotes that the current input requests the model to compute the detection loss. For this purpose, we adopt COCO dataset~\citep{lin2014microsoft} for training. In addition, to augment the model's discriminative capacity for diverse categories, we employ all category names from the training dataset as the textual input for Grounding DINO.

\paragraph{Referring expression comprehension data.} 
REC data typically provides images and the descriptive phrase corresponding to the target bounding box. We formulate questions and answers with a subsequent template:
\emph{“\textbf{User:} }\texttt{<image>} \emph{What is }\{caption\} \emph{in this image? Please output object location. \textbf{Assistant:} It is }\texttt{<DET>}\emph{.”}
where the \{caption\} is the descriptive phrase supplied in the dataset, generally a noun description augmented with several adjectives. To guarantee the diversity of the training data, we have also devised a variety of similar question templates, which are randomly chosen during the training process. We adopt most widely used REC dataset RefCOCO~\citep{kazemzadeh2014refcoco}, RefCOCO+~\citep{kazemzadeh2014refcoco}, RefCOCOg~\citep{mao2016refcocog}.

\paragraph{Reasoning detection data.} 
To enhance the model's applicability in scenarios necessitating comprehension of reasoning questions, such as embodied AI robots, we processed the ReasonSeg dataset~\citep{lai2023lisa} at the instance level to yield a reasoning detection benchmark called \textbf{ReasonDet}. Similar to the partition of ReasonSeg data from LISA~\citep{lai2023lisa}, the training set comprises 239 images and 1326 texts, while the validation set includes 200 images and 344 texts. During training, the questions are divided by their lengths. The template for short questions aligns with the REC template. For long questions, we adhere to the following template:
\emph{“\textbf{User:} }\texttt{<image>} \{question\} \emph{Please output object location and explain the reason. \textbf{Assistant:} Sure, the detection result is }\texttt{<DET>}\emph{, }\{reason\}\emph{.”}
where \{question\} is a long question asked in a natural language scenario, and \{reason\} is the explanation given by the model for the current box prediction.

\paragraph{Visual question answering data.} 
To preserve the inherent visual question answering (VQA) capability of the multimodal large language model, we also incorporated the LLaVA-Instruct-150k~\citep{liu2023llava} dataset generated by GPT-4 during training.

\section{Experiment}
\label{sec:exp}
\begin{table}[t]
\centering
\caption{Comparison of training costs across various models.}
\begin{tabular}{cc}
\toprule
Model      & GPU days \\ \midrule
DQ-DETR~\citep{liu2023dq}    & 70       \\
Shikra~\citep{chen2023shikra}     & 40       \\
MiniGPT-v2~\citep{chen2023minigpt} & 35     \\
Lenna      & \textbf{7}       \\ \bottomrule
\end{tabular}
\label{tab:time}
\end{table}
\begin{figure*}[t]
  \centering
   \includegraphics[width=0.97\linewidth]{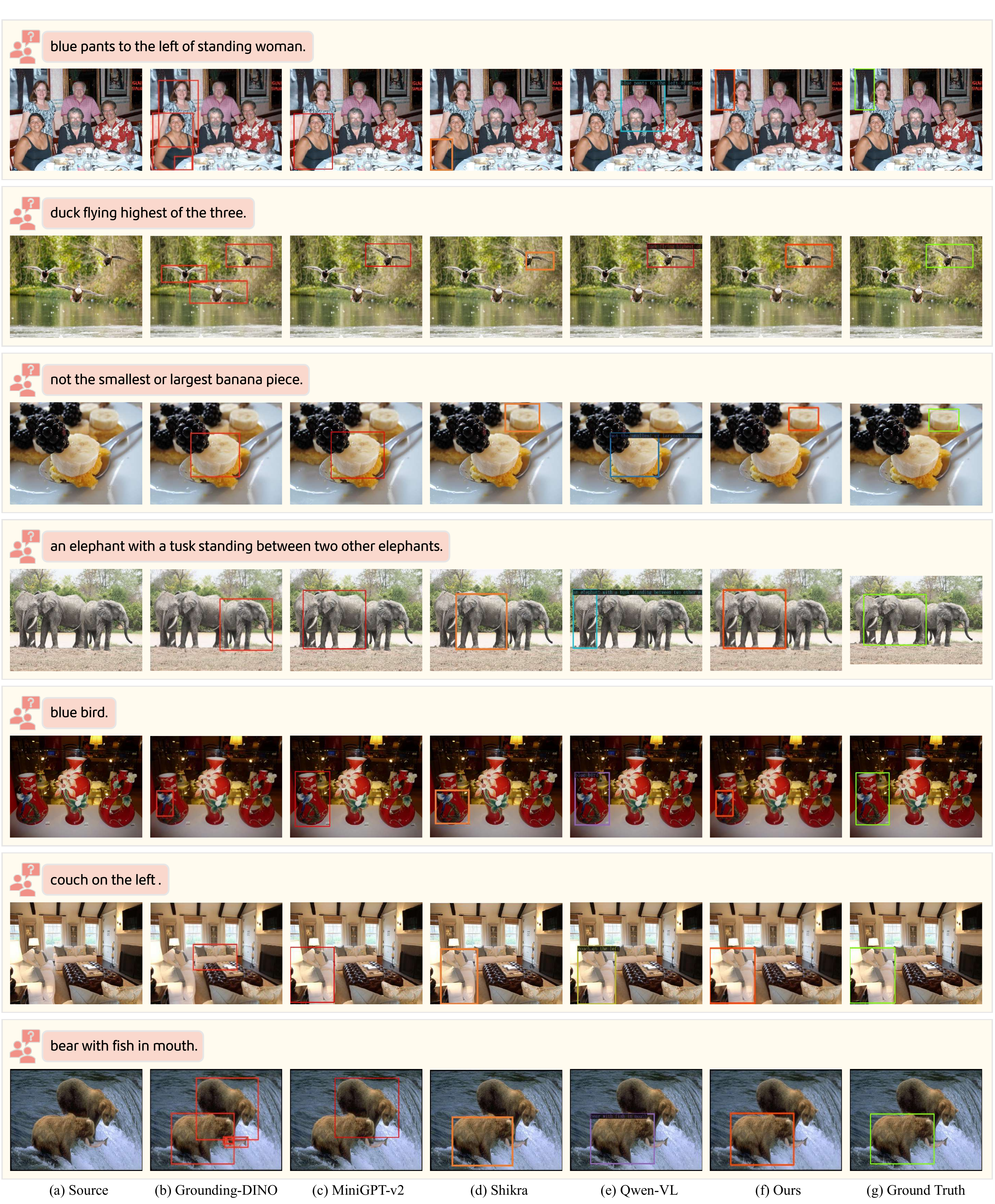}
   \caption{Visualization comparison with other VLMs on various referring expressions. }
   \label{fig:compare}
\end{figure*}

\begin{table*}[]
\newcommand{\tabincell}[2]{\begin{tabular}{@{}#1@{}}#2\end{tabular}}
\footnotesize
\renewcommand{\arraystretch}{1.5}
\setlength\tabcolsep{7pt}
\centering
\caption{Results on referring expression comprehension task. The best results are marked in bold.}
\begin{tabular}{cccccccccc}
\toprule
\multirow{2}*{Type}                                    & \multirow{2}*{Model}  & \multicolumn{3}{c}{RefCOCO} & \multicolumn{3}{c}{RefCOCO+} & \multicolumn{2}{c}{RefCOCOg} \\
                                                         &                         & val     & testA   & testB   & val      & testA   & testB   & val           & test         \\ \midrule
\multirow{11}{*}{\tabincell{c}{Specialist SOTAs \\ (Specialist/Finetuned)}} 
& TransVG~\citep{deng2021transvg} & 81.02 & 82.72 & 78.35 & 64.82 & 70.70 & 56.94 & 68.67 & 67.73 \\
& UNITER~\citep{chen2020uniter}  & 81.41 & 87.04 & 74.17 & 75.90 & 81.45 & 66.70 & 74.86 & 75.77 \\
& VILLA~\citep{gan2020large} & 82.39 & 87.48 & 74.84 & 76.17 & 81.54 & 66.84 & 76.18 & 76.71 \\
& RefTR~\citep{li2021referring}   & 85.65 & 88.73 & 81.16 & 77.55 & 82.26 & 68.99 & 79.25 & 80.01 \\
& MDETR~\citep{kamath2021mdetr}     & 86.75 & 89.58 & 81.41 & 79.52 & 84.09 & 70.62 & 81.64 & 80.89 \\
& UNICORN~\citep{yan2022towards} & 88.29 & 90.42 & 83.06 & 80.30 & 85.05 & 71.88 & 83.44 & 83.93 \\
& DQ-DETR~\citep{liu2023dq}           & 88.63   & 91.04   & 83.51   & 81.66    & 86.15   & 73.21   & 82.76         & 83.44        \\
& InstructDet~\citep{dang2023instructdet}  & 88.92   & 90.86   & 85.57   & 78.27    & 83.39   & 71.04   & 83.01         & 82.91        \\ 
& Grounding-DINO~\citep{koh2023grounding}                & 89.19   & 91.86   & 85.00   & 81.09    & 87.40   & 74.71   & 84.15         & 84.94        \\
\midrule
\multirow{5}*{\tabincell{c}{Generalist VL SOTAs \\ (w/o finetuning)}}    
& OFA-L~\citep{wang2022ofa}    & 79.96 & 83.67 & 76.39 & 68.29 & 76.00 & 61.75 & 67.57 & 67.58 \\
& VisionLLM-H~\citep{wang2023visionllm} & - & 86.70 & - & - & - & - & - & - \\
& Shikra-13B~\citep{chen2023shikra}                  & 87.83   & 91.11   & 81.81   & 82.89    & 87.79   & 74.41   & 82.64         & 83.16        \\
& MiniGPT-v2-7B~\citep{chen2023minigpt}               & 88.69   & 91.65   & 85.33   & 79.97    & 85.12   & 74.45   & 84.44         & 84.66        \\
& PerceptionGPT-13B~\citep{pi2023perceptiongpt} & 89.17 & 93.20 & 85.96 & 83.72 & 89.19 & 75.31 & 83.75 & 84.69 \\
& Qwen-VL-7B~\citep{Qwen-VL}                 & 89.36   & 92.26   & 85.34   & 83.12    & 88.25   & 77.21   & 85.58         & 85.48        \\
\midrule
\multirow{1}*{\tabincell{c}{ours}} & Lenna-7B    & \textbf{90.28}   & \textbf{93.22}   & \textbf{86.97}   & \textbf{88.08}    & \textbf{90.07}   & \textbf{83.99}   & \textbf{90.30}        & \textbf{90.29}        \\
                            
                                                         \bottomrule
\end{tabular}
\label{tab:rec}
\end{table*}
 
\begin{table}[]
\centering
\caption{Quantitative results on the validation set of ReasonDet with the metric of accuracy. RD refers to the training set of ReasonDet.}
\begin{tabular}{cc}
\toprule
Exp                    & \emph{Acc.} \\ 
\midrule
Qwen-VL~\citep{Qwen-VL}                & 12.96 \\
Shikra~\citep{chen2023shikra}              & 20.27 \\
MiniGPT-v2~\citep{chen2023minigpt}                & 25.25 \\
\midrule
Lenna (w/o RD) & 37.21     \\
Lenna               & \textbf{46.84}     \\ 

\bottomrule
\end{tabular}
\label{tab:reason}
\end{table}
\begin{figure*}[t]
  \centering
   \includegraphics[width=0.97\linewidth]{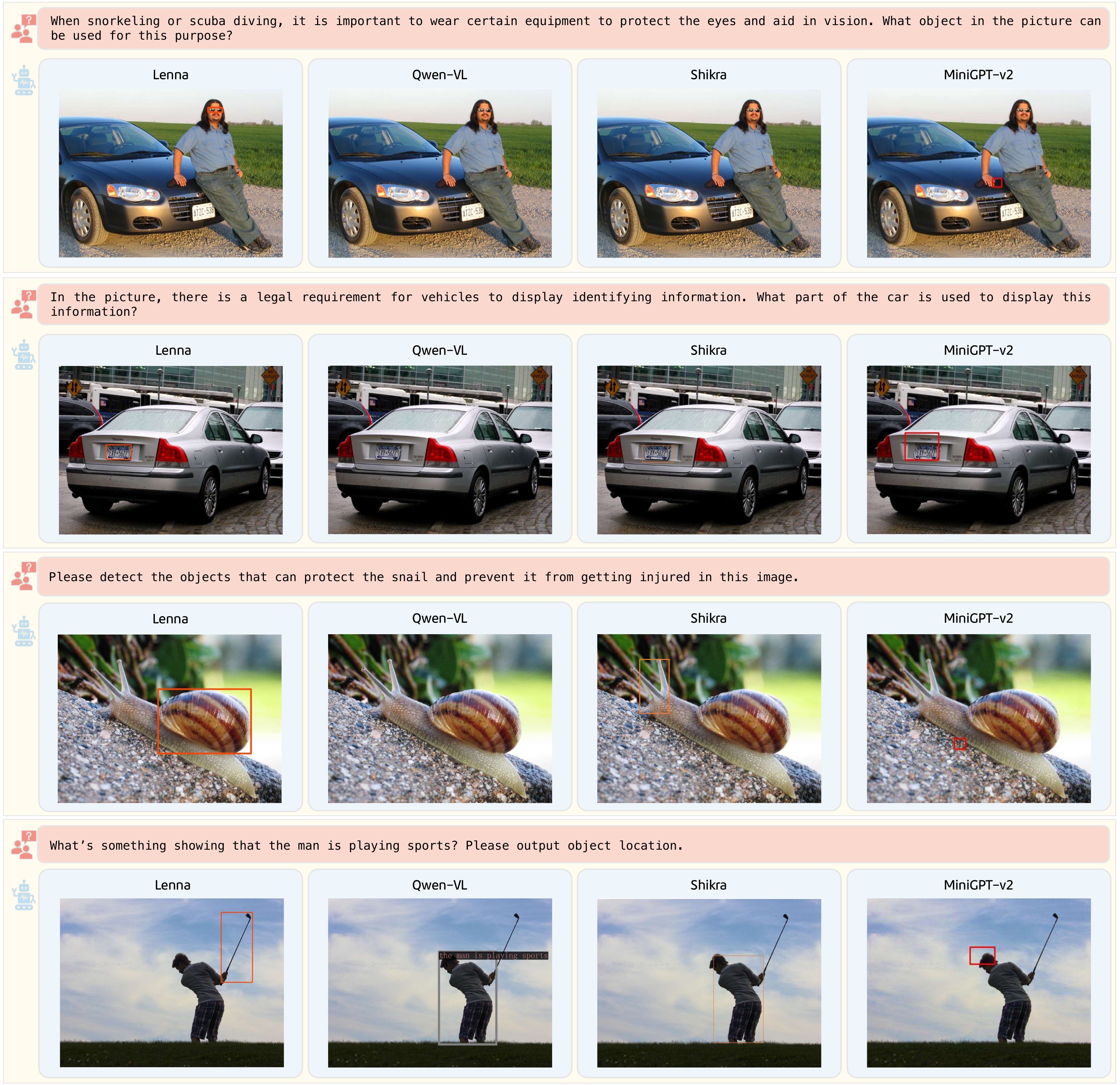}
   \caption{Visualization comparison of reasoning capability with generalist VLMs.}
   \label{fig:compare_reason}
\end{figure*}



\begin{table}[]
\setlength{\tabcolsep}{2pt}
\centering
\caption{Comparison on pre-trained parameters. The second column is the model chosen for detector $\mathcal{D}$. We evaluate through the metric of accuracy on the val set of the REC dataset. }
\begin{tabular}{ccccc}
\toprule
{Exp}  & {$\mathcal{D}$} & {RefCOCO} & {RefCOCO+} & {RefCOCOg} \\
\midrule
E1            & From Scratch & 24.46       & 23.74           & 33.48               \\
E2            & GDINO-T   & 89.27       & 87.47           & 88.50               \\
E3            & GDINO-B   & 89.36         & 85.37      & \textbf{90.57}         \\
E4            & GDINO-L   & \textbf{90.28}      & \textbf{88.08}      & 90.30    \\
\bottomrule
\end{tabular}
\label{tab:param}
\end{table}



\begin{table}[]
\setlength{\tabcolsep}{2pt}
\centering
\caption{Ablation study on training data. We evaluate through the metric of accuracy on the val set of the REC dataset. }
\begin{tabular}{ccccccc}
\toprule
\multirow{2}{*}{Exp} & \multicolumn{3}{c}{Dataset} & \multirow{2}{*}{RefCOCO} & \multirow{2}{*}{RefCOCO+} & \multirow{2}{*}{RefCOCOg} \\
                       & REC       & OD        & VQA        \\ \midrule
E1                     & \Checkmark &                           &                           & 31.78        & 30.85      & 20.53               \\
E2                     & \Checkmark &                           & \Checkmark & 85.62        & 85.35      & 87.01          \\
E3                     & \Checkmark & \Checkmark &                           & 89.45      & 85.92        & 88.77          \\

E4                     & \Checkmark & \Checkmark & \Checkmark & \textbf{90.28}     & \textbf{88.08}   & \textbf{90.30}     \\ \bottomrule
\end{tabular}
\label{tab:traindata}
\end{table}
\subsection{Implementation Details}
\paragraph{Network architecture.}
Lenna consists of a multimodal large language model $\mathcal{M}$ and an open-set detector $\mathcal{D}$. Within the multimodal large language model, we employ pre-trained LLaVA-7B~\citep{liu2023llava} for efficient training. Our open-set detector is designed on Grounding-DINO with Swin-Large Transformers~\citep{liu2021swin} as the vision backbone.
During the training stage, LoRA~\citep{hu2021lora} is utilized for an efficient finetuning of LLM. We thoroughly train the MQS module and decoder in the detector, while all other parameters are frozen to preserve the original capabilities of the pre-trained model.
\paragraph{Hyperparameter.}
The training takes approximately 20 hours on 8 NVIDIA A100 GPUs. We adopt AdamW~\citep{loshchilov2017decoupled} optimizer with a learning rate of 3e-4 and use a learning rate scheduler WarmupDecayLR with the warmup steps of 10. 
The batch size is set to 2 on each device. The modulating parameters $\lambda_{tok}$ and $\lambda_{det}$ in total loss $\mathcal{L}$ are set to 1.0 and 1.0, respectively. In $\lambda_{det}$, there are $\lambda_{L1}$, $\lambda_{GIOU}$ and $\lambda_{Contrast}$ which are set to 5.0, 2.0 and 1.0.

\subsection{Comparison with State-of-the-art Methods}
\paragraph{Training resource consumption.}
As shown in Table~\ref{tab:time}, we compare the training resource consumption with DQ-DETR~\citep{liu2023dq}, Shikra~\citep{chen2023shikra} and MiniGPT-v2~\citep{chen2023minigpt} on NVIDA A100 GPUs. In contrast to existing works, Lenna incurs a significantly lower training cost with its simplistic model architecture and efficient training strategy.

\paragraph{Quantitative comparison.}
To guarantee the impartiality of the comparison, we evaluate all methods on RefCOCO, RefCOCO+, and RefCOCOg using the accuracy metric with an IoU of 0.5. As shown in Table~\ref{tab:rec}, we first compare our method with specialists that encompass localization models and fine-tuned generalist or foundation models on localization tasks, such as TransVG~\citep{deng2021transvg}, UNITER~\citep{chen2020uniter}, VILLA~\citep{gan2020large}, RefTR~\citep{li2021referring}, MDETR~\citep{kamath2021mdetr}, UNICORN~\citep{yan2022towards}, DQ-DETR~\citep{liu2023dq}, InstructDet~\citep{dang2023instructdet}, GroundingDINO~\citep{koh2023grounding}. Additionally, our method is also compared with generalist VL models without fine-tuning, including  OFA~\citep{wang2022ofa}, VisionLLM~\citep{wang2023visionllm}, Shikra~\citep{chen2023shikra}, MiniGPT-v2~\citep{chen2023minigpt}, PerceptionGPT~\citep{pi2023perceptiongpt}, Qwen-VL~\citep{Qwen-VL}. These generalist VL models are capable of undertaking a variety of vision-language tasks, including image captioning, VQA, REC, \etc. 
Among all listed models, Lenna comprehensively demonstrates a promising performance advantage.

The quantitative results on the ReasonDet dataset are presented in Table~\ref{tab:reason}. We performed a comparative analysis between our proposed method, Lenna, and existing MLLM methods, including MiniGPT-v2~\citep{chen2023minigpt}, Shikra~\citep{chen2023shikra}, and Qwen-VL~\citep{Qwen-VL}. To ensure a fair comparison, we excluded the ReasonDet data from the training dataset, as indicated in the Lenna (w/o RD) row in Table~\ref{tab:reason}. The results demonstrate that irrespective of whether ReasonDet data is used in training, our method significantly surpasses other techniques. Lenna (w/o RD) achieves a 47.37\% improvement over the best-performing MiniGPT-v2 in the SOTAs, and Lenna even exceeds this by 85.50\%. This provides solid evidence that Lenna can genuinely comprehend the content within the problem and accomplish precise positioning.

\paragraph{Qualitative comparison.}
Figure~\ref{fig:compare} depicts a qualitative comparison, showcasing an array of scenes and REC results produced by diverse methods. The results suggest that while Grounding-DINO is adept at comprehending simple and explicit text information such as color and position, it struggles with scenarios where understanding the relationship between multiple target positions within an image is required. A common characteristic of MiniGPT-v2~\citep{chen2023minigpt}, Shikra~\citep{chen2023shikra}, and Qwen-VL~\citep{Qwen-VL} is that the discrete output form of their language models results in an increased challenge in object location and a certain degree of loss in positioning accuracy. 
On the other hand, our method consistently outperforms in understanding complex language information and achieving accurate localization.

Figure~\ref{fig:compare_reason} further illustrates the comparison in reasoning capability. Most existing methods struggle with this task, as the reasoning task diverges from the REC task in that it requires the model to not only understand the question's real meaning but also possess a world knowledge base, which in turn enables reasoning and positioning. Figure~\ref{fig:compare_reason} showcases some results of MiniGPT-v2~\citep{chen2023minigpt}, Shikra~\citep{chen2023shikra}, Qwen-VL~\citep{Qwen-VL} and Lenna on the ReasonDet dataset, with Lenna demonstrating superior performance in various reasoning scenarios of differing difficulty levels, such as long-question (first and second rows) and short-question (third and fourth rows).
\subsection{Ablation Study}
\paragraph{Model scaling.}
To underscore the significance of the model scale, we have conducted a comprehensive series of comparative experiments on the scale of the detector. As depicted in E2-E4 in Table~\ref{tab:param}, the results suggest that a larger scale of $\mathcal{D}$ can effectively alleviate the model fitting complexity and confer superior performance enhancement on the model.
Furthermore, the pre-trained weights of the detector hold significant importance. As demonstrated in E1 of Table~\ref{tab:param}, training the detector from scratch results in a decline in model performance.

\paragraph{Training data.}
In Table~\ref{tab:traindata}, we display the contribution of each type of training data to the performance. It is evident that both object detection (OD) data and VQA data exert varying degrees of impact on the model performance. The OD data provides explicit guidance for semantic alignment to the model, while the VQA data contributes to the diversification of the \texttt{<DET>} embedding.
\section{Conclusion}
\label{sec:con}
We present Lenna, a novel framework that leverages the representational power and world knowledge of large language models (LLM) to enhance reasoning in object detection tasks. Lenna introduces a unique \texttt{<DET>} token embedding to facilitate accurate positioning without losing reasoning information.
Lenna stands out due to its efficient training and the ability to extend to various tasks with minimal additional costs. Its design simplicity allows for rapid adaptation and scaling, demonstrating a notable improvement over previous models in terms of training efficiency and versatility. 
Owing to Lenna's training efficiency and its expansive application potential, we aspire to furnish novel insights for future research and practical deployments in the domain of multimodal large language models.

\section{Acknowledgement}

This work is supported by National Key R\&D Program of China (No. 2022ZD0118700).
{
    \small
    \bibliographystyle{ieeenat_fullname}
    \bibliography{main}
}


\end{document}